\documentclass[runningheads]{llncs}

\usepackage{graphicx}
\usepackage{amssymb}
\usepackage{caption}
\usepackage{subcaption}
\usepackage{hyperref}
\usepackage{cite}
\usepackage{adjustbox}
\usepackage{multirow}
\usepackage{hhline}

\usepackage{booktabs}
\usepackage{lipsum}
\begin{document}
\title{SparseSwin: Swin Transformer with Sparse Transformer Block}
%

\author{
    Krisna Pinasthika \and
    Blessius Sheldo Putra Laksono \and
    Riyandi Banovbi Putera Irsal \and
    Syifa’ Hukma Shabiyya\and
    Novanto Yudistira
}

\authorrunning{K. Pinasthika et al.}
%
\institute{
    Faculty of Computer Science, Brawijaya University, Malang City, East Java, Indonesia, 65145 \\
    \email{\{krisnapinas,blessiussheldo,riyandiirsal1st,syifahukma\_s\}@student.ub.ac.id},
    \email{yudistira@ub.ac.id},
}

%
\maketitle              
\begin{abstract}
Advancements in computer vision research have put transformer architecture as the state-of-the-art in computer vision tasks. One of the known drawbacks of the transformer architecture is the high number of parameters, this can lead to a more complex and inefficient algorithm. This paper aims to reduce the number of parameters and in turn, made the transformer more efficient. We present the Sparse Transformer (SparTa) Block, a modified transformer block with an addition of a sparse token converter that reduces the number of tokens used. We use the SparTa Block inside the Swin-T architecture (SparseSwin) to leverage Swin’s capability to downsample its input and reduce the number of initial tokens to be calculated. The proposed SparseSwin model outperforms other state-of-the-art models in image classification with an accuracy of 86.96\%, 97.43\%, and 85.35\% on the ImageNet100, CIFAR10, and CIFAR100 datasets respectively. Despite its fewer parameters, the result highlights the potential of a transformer architecture using a sparse token converter with a limited number of tokens to optimize the use of the transformer and improve its performance. The code available at \url{https://github.com/KrisnaPinasthika/SparseSwin}

\end{abstract}

\section{Introduction}
\label{sec:introduction}
Computer vision has become one of the critical topics in the world of artificial intelligence. The majority of computer vision models use CNN as the method for image classification \cite{NIPS2012_c399862d, Simonyan15, DBLP:journals/corr/HeZRS15, DBLP:journals/corr/SzegedyLJSRAEVR14, DBLP:journals/corr/HowardZCKWWAA17, DBLP:journals/corr/SzegedyVISW15, DBLP:journals/corr/HuangLW16a, DBLP:journals/corr/abs-1905-11946}. When assessing the performance of the model, dataset such as Imagenet \cite{deng2009imagenet} is used because of its extensive data and variety of classes. Using large data in the model’s training phase dramatically improves performance when tested on real-life or new data. The use of CNN on many image classification models made CNN evolve to a more sophisticated scale, connectivity, and forms of convolution \cite{chen2021review}.

However, the evolution of network topologies in Natural Language Processing (NLP) has taken a different path, with Transformer now becoming the dominant architecture \cite{vaswani2017attention}. Transformer dominance has even entered the subject of computer vision with the introduction of the Vision Transformer \cite{dosovitskiy2020image}. Vision Transformer modifies the transformer model for Natural Language Processing (NLP) for image classification. Vision Transformer’s architecture uses a stacked Transformer Block to process image patches that do not overlap \cite{zhang2023vision}. Compared to its rival CNN, Transformer-based models have large receptive fields \cite{cheng2023hybrid} and superior performance on large datasets \cite{liu2021swin}. Despite the outstanding performance, excessive attention creates high computing costs because of the dependence on the image input size in carrying out self-attention calculations, and it takes a long time to converge while training.

There is various state-of-the-art to use Transformers in image classification problems which enhances the original Vision Transformer model and makes it more robust and efficient. One of the state-of-the-art research on transformers for image classification is the Swin Transformer. Swin Transformer avoids the excessive use of attention on the whole image by using local window-based attention \cite{liu2021swin} and manages to reduce the computing costs of using a transformer on images. On the other hand, SparseFormer \cite{gao2023sparseformer} adjusts the region of interest (RoI) by focusing on the front of the image using several latent tokens. This was done to extract features from the image. The advantage of this approach is that the calculation is done on a limited number of tokens, so the calculation of self-attention is not affected by the size of the input image but depends on the size of the latent token thus improving efficiency.

In previous state-of-the-art research, the use of a transformer for image classifications has a problem with a large number of parameters. For example, the original Vision Transformer adopts a general transformer model formerly used in NLP resulting in 85M parameter. While other research such as \cite{liu2021swin} \cite{gao2023sparseformer} manages to reduce the number of parameters to 27.6 M and 32M it still pales in comparison with other small and lightweight CNN-based models such as \cite{DBLP:journals/corr/abs-1905-11946, DBLP:journals/corr/HeZRS15} with 12M and 26M parameters.

Inspired by the success of Swin Transformer \cite{liu2021swin} and SparseFormer \cite{gao2023sparseformer}, we combine local attention with the windowing technique used by Swin Transformer and the use of a limited latent token to improve efficiency further. In our approach, we employ the Swin Transformer to process the image in several stages and feed it into Sparse Transformer (SparTa) block, which compresses the amount of token and provides results that can be classified in the classification layer. Since this Transformer Block uses a limited number of tokens in performing computation, it is reasonable to call this block Sparse as done by the research conducted by \cite{gao2023sparseformer}. The computation performed on SparTa Block depends on the token size so it is smaller when compared to the input image.

The proposed SparseSwin model overtook several other models in image classification. SparseSwin model has a smaller number of parameters, 17.58M when compared to the 27.6M parameter in the Swin-T model and 85M parameter in the ViT-B model. The reduction in the number of parameters in SparseSwin is also followed by an improvement in accuracy when compared to state-of-the-art research with identical parameters. SparseSwin achieved accuracies of 86.96\%, 97.43\%, and 85.35\% on the ImageNet100, on CIFAR10, and CIFAR100 datasets respectively. In summary, our contributions are:
\begin{enumerate}
  \item We propose a Sparse Transformer (SparTa) Block architecture using limited latent tokens to improve the efficiency and performance of the Swin Transformer for image classification.
  \item We examined the effect of regularization on the attention weight obtained in SparTa Block.
  \item We obtained accuracy improvement on ImageNet100, CIFAR10, and CIFAR100 benchmark datasets.
  \item We obtained a smaller parameter size compared to the state-of-the-art Transformer model.
\end{enumerate}

The paper is divided into 5 sections: Section \ref{sec:related_works} provides an overview of relevant previous research. Section \ref{sec:proposed_method} outlines our proposed approach. Section \ref{sec:experiments} shows the experimental results, which assess the effectiveness of our model. Section \ref{sec:conclusion} provides a summary of our contributions.

\section{Related Works}
\label{sec:related_works}

\textbf{Transformer Attention Mechanism}. The Attention mechanism is adapted from how humans are selective about the information \cite{gong2022swin}. This mechanism allows the model to focus more on the important part of the input to produce accurate output. The Attention Mechanism initially introduce through the Recurrent Attention Model (RAM) \cite{mnih2014recurrent}. Attention Mechanism continues to develop to create various variations such as hard attention \cite{xu2015show}, self-attention \cite{vaswani2017attention}, multi-head attention \cite{vaswani2017attention}, cross-modal attention \cite{lu2018neural}, and attention with Convolutional Neural Network (CNN) \cite{woo2018cbam}. Then, in recent research, an attention mechanism was developed to pay more attention to certain areas of the image and ignore less important areas called the Focusing Transformer implemented in SparseFormer \cite{gao2023sparseformer}. This research will utilize a modified version of the Focusing Transformer as a method for retrieving important features with input generated from the output of the Swin Transformer Block.

\textbf{Vision Transformer}. Vision Transformer (ViT) \cite{dosovitskiy2020image} was developed based on the Transformer concept applied to NLP, which is then applied to image recognition. ViT will split the image into a set of patches of the same size and be considered as a sequence. The sequence of patches will be used as input for the Transformer model. ViT becomes the inspiration of many of the further works, one of which uses shifting windows to reduce the complexity of the algorithm \cite{gong2022swin} ViT achieves an accuracy of up to 88\% when trained on the ImageNet1K dataset. This result is better when compared to other CNN-based solutions, such as ResNet152, with an accuracy of 87\%.

\textbf{Swin Transformer}. The development of methods for vision tasks is extensive. In the previous Vision Transformer \cite{dosovitskiy2020image}, the resulting computational complexity increases quadratically with the input image size due to the global attention calculation. To overcome this, a backbone algorithm is recently proposed in the Transformer that has a linear computational complexity calculation for the image size. The Swin Transformer will start with small patches and then, in deeper layers of the Transformer, will merge adjacent patches \cite{liu2021swin}. The Swin Transformer scheme is a shifted window that is flexible. Swin Transformer achieved an accuracy of 84\%, which is 5\% better compared to the base ViT model from previous research. This research utilizes the adaptation of the Swin Transformer with the addition of a Focusing Transformer on the resulting output.

\textbf{SparseFormer}. This paper proposes a new model for using transformer architecture in solving computer vision tasks. SparseFormer uses a limited number of latent tokens used in customizing Region of Interest (RoI) tokens by focusing on the front of the image with a limited number of tokens to extract features from the image. Since the computation is only performed on a limited number of tokens, the amount of computation performed does not depend on the size of the input image so that the computation process can be more efficient \cite{gao2023sparseformer}.

\section{Proposed Method}
\label{sec:proposed_method}
\begin{figure}[bp]
    \centering
    \includegraphics[width=1\textwidth]{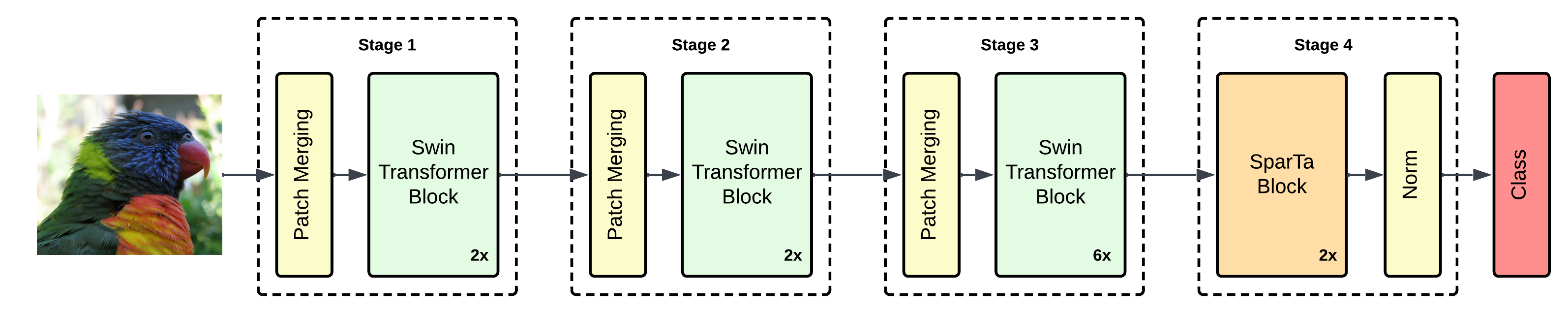}
    \caption{The architecture of SparseSwin}
    \label{fig:architecture}
\end{figure}

In this section, we describe in detail the architecture of SparseSwin. SparseSwin is constructed from the combination of the Swin Transformer and Sparse Transformer Block (SparTa Block). This architecture consists of 4 stages with stages 1 to 3 using the same Patch Merging Block and Swin Transformer Blocks as the Swin Transformer \cite{liu2021swin}, Whereas in stage 4 we use our proposed SparTa Block followed by a Layer Normalization \cite{ba2016layer}. Specifically, our proposed architecture is illustrated in Fig. \ref{fig:architecture}.

Initially, we propagate the input image to “Stage 1”. In this stage, Patch Merging is applied to split the input image into several non-overlapping patches with a patch size of 4 which is then followed by applying linear embedding as performed by \cite{liu2021swin}. Patch merging is used to reduce the size of the feature map while maintaining the spatial information with an output size of $\frac{H}{8}\times\frac{W}{8}$. The same is done in "Stage 2" and "Stage 3" with output sizes of $\frac{H}{16}\times\frac{W}{16}$ and $\frac{H}{32}\times\frac{W}{32}$. Thereafter, the feature map obtained from Patch Merging will be propagated to the Swin Transformer Blocks. The Swin Transformer Blocks used in this research are the same as the previous research \cite{liu2021swin}. 

\begin{figure}[htp]
    \centering
    \includegraphics[width=1\textwidth]{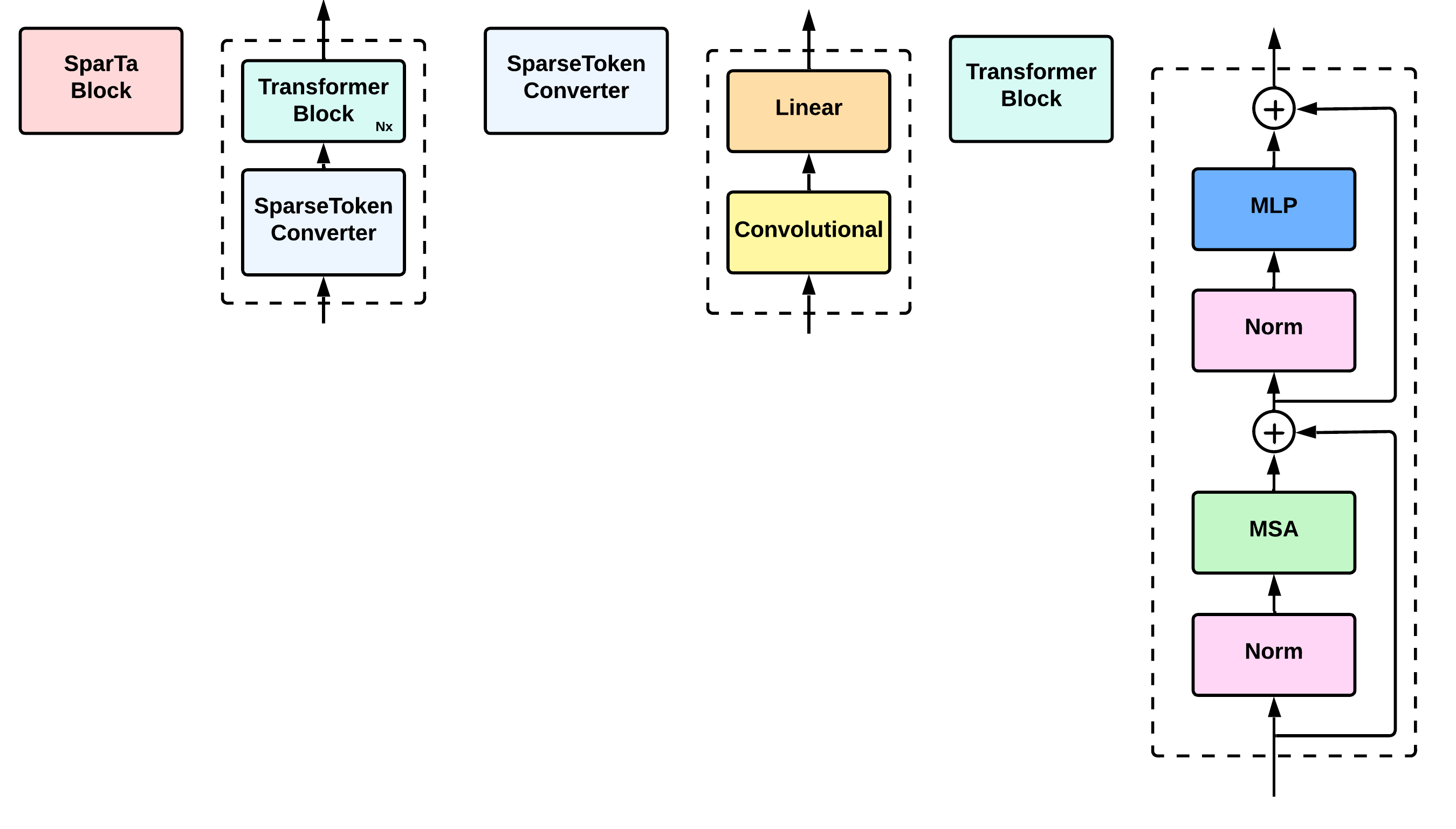}
    \caption{The successive SparTa blocks in Stage 4 of SparseSwin for image classification}
    \label{fig:sparta_successive_blocks}
\end{figure}

Following the feature map obtained in Stage 3, the feature map will be propagated to the SparTa Block. SparTa Block consists of two main parts: the Sparse Token Converter and the regular Transformer Block. In detail, the SparTa Block architecture is illustrated in Fig. \ref{fig:sparta_successive_blocks}. SparTa Block will change the representations of features generated by the previous Stages \textit{B}, \textit{C}, \textit{H}, and \textit{W} which are Batch size, Height, Channel, and Width respectively into token sizes along with embedding with an output size of \textit{B}, \textit{t}, \textit{e} which is the Batch size, Number of latent tokens, and Size of the embedding respectively. The process that occurs within the SparTa Block will be described in more detail in the next section. The output generated from the SparTa Block will be propagated into the normalization layer \cite{ba2016layer}. The normalization Layer changes the input data range from different scales to the same range to prevent the accumulation of large error gradients that cause the training to be unstable.

\textbf{Sparse Token Converter}. Sparse Token Converter is used to convert the dimension size of the output of the previous stage into a new token with a new embedding dimension size. The conversion process is done using a convolution layer \cite{lecun1995convolutional} to acquire the new embedding and a linear layer to convert the size of the patches into latent tokens. Specifically, the kernel and stride sizes used in the convolution layer are 3 and 1. Meanwhile, the input features in the linear layer are $\textit{H}\times \textit{W}$ in stage 3 with an output feature of 49 to produce a new token representation. The output obtained from the Sparse Token Converter ($S_T$) is $S_T\in\ \mathbb{R}^{t\ \times\ e}$ where \textit{t} is the desired number of new tokens and \textit{e} is the number of new embedding dimension.

\textbf{Regular Transformer Block}. The regular Transformer Block used in this research is the same transformer block as the Transformer Block introduced by \cite{vaswani2017attention}. The MSA calculation is done by receiving $S_T$ input that has been normalized with Layer Normalization (LN) \cite{ba2016layer}. Furthermore, the normalized MSA output will be propagated to the Multilayer Perceptron (MLP) with a hidden ratio size of 4. Regular transformer block calculation is presented in Eqs \ref{eq:eq1} and \ref{eq:eq2} where $\hat{z}^l$  and $z^l$ are the output features of the MSA block and MLP block. Specifically, in this research, the loops used in the Regular Transformer Block used are 2.

\begin{equation}
    \hat{z}^l = MSA(LN(S_T))+ S_T 
    \label{eq:eq1}
\end{equation}

\begin{equation}
    z^l = MLP(LN(\hat{z}^l))+ \hat{z}^l 
    \label{eq:eq2}
\end{equation}

Therefore, the overall output of SparTa Block is $Block_{S_T}$ $\in$ $\mathbb{R}^{b \times t \times e}$  where $b$ is the size of the batch used. The computation of MSA using SparTa Block is not affected by the size of the input image. However, it depends on the number of tokens used, so in general, the use of SparTa Block can reduce the size of ordinary MSA computation.

\section{Experiments}
\label{sec:experiments}
We conducted several experiments to observe SparseSwin's performance using ImageNet100 \cite{russakovsky2015imagenet}, CIFAR10 \cite{NIPS2012_c399862d}, and CIFAR100 \cite{NIPS2012_c399862d} image classification datasets. We also experiment with regularization of the attention weights to prevent the model from overfitting and make the model generalize better. Furthermore, we will compare the best results obtained in the previous experiments with other state-of-the-art research.

\subsection{Implementation Details}
\label{sec:Implementation_Details}
\textbf{Configuration on ImageNet100}. In the model training process, we used Adam \cite{kingma2014adam} as the optimizer with a learning rate of 1e-4 and batch size of 128. The model is trained on ImageNet100, and this dataset is a subset of the ImageNet1K dataset with only 100 classes. The training process is carried out using NVIDIA A100. The model is trained up to 100 epochs to find the point where the training process converges based on the change of loss value. In the training phase of SparseSwin, we freeze “Stage 1” and “Stage 2” to prevent the training process on low-level features. The freeze is done because the backbone has trained on ImageNet1K, so the training is only devoted to higher-level features. In detail, the training process is carried out using a total of 49 tokens with an embedding size and num MSA heads in the SparTa Block of 512 and 16 respectively. The number of SparTa Block loops used is 2 and uses the QKV bias.

\textbf{Configuration on CIFAR}. We fine-tune the model with the AdamW optimizer \cite{loshchilov2017decoupled}, which proved to be successful in improving its performance. This optimizer has a learning rate of 1e-5 and a weight decay of 0.01, which helps to adjust the model's parameters. The input data is resized to 224 dimensions to match the model to the SparseSwin architecture. Furthermore, data augmentation techniques such as random cropping, random horizontal flips, and resizing cropping are used to expand and diversify the dataset, boosting the model's capacity to generalize and handle different scenarios. The detailed architecture used for training on the CIFAR10 and CIFAR100 datasets uses the same configuration as ImageNet100 using the NVIDIA A100.

\subsection{The Influence of The Sparsity Attention Weight in Sparta Block for Image Classification}
\label{sec:influence_sparsity}
In this research, we conduct further experiments to test SparseSwin's ability to perform image classification. We experiment by testing the effectiveness of L1 \cite{tibshirani1996regression} and L2 \cite{golub1999tikhonov} regularization on the attention weight obtained by SparTa Block in the training phase. Attention weight was chosen in this study because it was inspired by the research conducted by Zhang et al. \cite{zhang2018attention} related to the case of neural machine translation and abstractive summarization, that the attention mechanism is used to guide the prediction process sequentially focusing on specific words thus the higher the attention weight, the more the corresponding word contribution. Nevertheless, not all words are relevant to produce a certain output. Thus, we apply the problem to the case of image classification by making the attention weight sparser, allowing SparTa Block to select more relevant features in the case of image classification. Besides making the attention weight sparser, the use of regularization has been proven to prevent overfitting and can make the model generalize better. In this study, we tested the lambda values used for regularization including 1e-4 and 1e-5. The experimental results are reported in Table \ref{tab:training_withl2_history}.

Based on the results of our evaluation metrics testing, SparseSwin has been able to provide good evaluation results with 86.64\%, 97.43\%, and 85.35\% accuracy on ImageNet100, CIFAR10, and CIFAR100 datasets, respectively. The test continued by training SparseSwin with L1 and L2 regulators on the attention weights obtained in the SparTa Block. Then, we continued testing the trained model and found that the use of regularization gave slightly better results on the ImageNet100 dataset with 86.96\% accuracy while it showed no improvement on the CIFAR10 and CIFAR100 datasets. Based on our analysis, L1 and L2 regularization did not have a significant impact on the CIFAR10 and CIFAR100 datasets because the number of datasets used was not as large as the ImageNet100 dataset. Additionally, considering the fact that we increased the resolution of the input images on the CIFAR10 and CIFAR100 datasets from $32^2$ to $224^2$ could potentially hinder the effectiveness of the regularization.

\begin{table}[htp]
\extracolsep{\fill}
\def\arraystretch{1.25}%
\centering
\large
\caption{Quantitative comparison of SparseSwin using L1 \& L2 Regularizer on Imagenet100, CIFAR10, and CIFAR100}
\label{tab:training_withl2_history}
\resizebox{\columnwidth}{!} {
    \begin{tabular}{c | c | c | c | c | c}
    \hline
    \multirow{2}{*}{Model} & \multirow{2}{*}{Lambda Regularizer} & \multirow{2}{*}{Input Resolution} & \multicolumn{3}{c}{Accuracy(\%)} \\
    \hhline{~~~---} &    &    & ImageNet100 & CIFAR10 & CIFAR100 \\
    \hline
    SparseSwin & - & $224^2$ & 86.64 & \textbf{97.43} & \textbf{85.35} \\
    SparseSwin with L1 & 1e-4 & $224^2$ & 86.68 & 97.42 & 84.70 \\
    SparseSwin with L1 & 1e-5 & $224^2$ & 86.88 & 97.28 & 84.86 \\
    SparseSwin with L2 & 1e-4 & $224^2$ & \textbf{86.96} & 97.31 & 84.80 \\
    SparseSwin with L2 & 1e-5 & $224^2$ & 86.90 & 97.31 & 84.88\\
    \hline
    \end{tabular}
}
\end{table}

\subsection{Results on ImageNet100}
\label{sec:imageNet100}

We conducted a performance comparison between SparseSwin and other state-of-the-art models using the ImageNet100 dataset as a benchmark. We found several types of models with the same input resolution of $224^2$ with various parameter sizes. The experimental results are presented in Table \ref{tab:imagenet100}. Based on the experimental results that have been carried out, SparseSwin achieves the smallest number of parameters of $~17$ M when compared to Swin-T \cite{liu2021swin} and ViT-B \cite{dosovitskiy2020image}  models with $~27$ M and $~85$ M parameters. Despite having a much smaller number of parameters than other transformer models, it is obvious that the SparseSwin model achieves higher accuracy than the previous study with an accuracy of 86.96\%. SparseSwin outperforms the recent state-of-the-art research and obtains higher accuracy with a difference of $\pm1.74\%$, $\pm6.06\%$, and $\pm7.66\%$ when compared to Swin-T, Vit-B, and DLME \cite{zang2022dlme} respectively.  

\begin{table}[bp]
\extracolsep{\fill}
\def\arraystretch{1.25}%
\centering
\large
\caption{Quantitative Comparison on ImageNet100}
\label{tab:imagenet100}
\resizebox{\columnwidth}{!}{
    \begin{tabular}{c c c c c}
        \hline
         Model & \#Params(100 class) & Input Resolution & Model Type & Accuracy(\%) \\
        \hline
        Swin-T \cite{liu2021swin} & 27.6 M & $224^2$ & Transformer & 85.22 \\
        ViT-B \cite{dosovitskiy2020image} & 85.87 M & $224^2$ & Transformer & 80.90 \\
        DLME(ResNet-50,linear) \cite{zang2022dlme} & $-$ & $224^2$ & Convolution & 79.3 \\
        \hline
        SparseSwin with L2 & 17.58 M & $224^2$ & Transformer & \textbf{86.96} \\
        \hline
    \end{tabular}
}
\end{table}

\subsection{Results on CIFAR10}
\label{sec:cifar10}
In this section, we compare the performance between SparseSwin and other state-of-the-art models using the CIFAR10 dataset as a benchmark. We found multiple model types with diverse input resolutions and parameter sizes, namely $32^2$ and $224^2$. Quantitative comparison between SparseSwin and state-of-the-art models is presented in Table \ref{tab:cifar10}. Compared to our proposed method, the input resolution size used in the DenseNet-BC-19 \cite{zhang2017mixup} and NesT-B \cite{zhang2022nested} models is $32^2$, which is smaller than the SparseSwin model. Despite having a smaller input resolution size, DenseNet-BC-19 and NesT-B produce a larger number of parameters compared SparseSwin, namely, 25.6 M and 97.2 M. With a larger number of parameters, DenseNet-BC-19 and NesT-B were still not performing optimally when compared to SparseSwin which achieved 97.43\% accuracy, with a margin of $\pm0.13\%$ and $\pm0.23\%$. 

Furthermore, there are various other state-of-the-art studies that have the same input resolution size as our proposed model including ResNet XnIDR \cite{sun2021xnodr}, CRATE-S \cite{yu2023white}, CRATE-B \cite{yu2023white}, and CRATE-L \cite{yu2023white} with the number of parameters 23.86 M, 13.12 M, 22.80 M, and 77.64 M respectively. Although SparseSwin has fewer parameters than state-of-the-art research, our proposed model outperformed other models by a slight improvement with a difference of $\pm0.56\%$, $\pm1.43\%$, $\pm0.63\%$, and $\pm0.23\%$ on ResNet XnIDR \cite{sun2021xnodr}, CRATE-S \cite{yu2023white}, CRATE-B \cite{yu2023white}, and CRATE-L \cite{yu2023white} respectively. 

\begin{table}[tp]
\extracolsep{\fill}
\def\arraystretch{1.25}%
\centering
\large
\caption{Quantitative Comparison on CIFAR10}
\label{tab:cifar10}
\resizebox{\columnwidth}{!}{
\begin{tabular}{c c c c c}
    \hline
     Model & \#Params(10 class) & Input Resolution & Model Type & Accuracy(\%) \\
    \hline
    DenseNet-BC-190+Mixup\cite{zhang2017mixup} & 25.6 M & $224^2$ & Convolution & 97.3\\
    ResNet XnIDR \cite{sun2021xnodr} & 23.86 M & $224^2$ & Convolution & 96.87 \\
    NesT-B\cite{zhang2022nested} & 97.2 M & $32^2$ & Transformer & 97.2 \\
    CRATE-S\cite{yu2023white} & 13.12 M & $224^2$ & Transformer & 96 \\
    CRATE-B\cite{yu2023white} & 22.80 M & $224^2$ & Transformer & 96.8 \\
    CRATE-L\cite{yu2023white} & 77.64 M & $224^2$ & Transformer & 97.2 \\
    \hline
    SparseSwin & 17.58 M & $224^2$ & Transformer & \textbf{97.43 }\\
    \hline
\end{tabular}
}
\end{table}

\subsection{Results on CIFAR100}
\label{sec:cifar100}
In this section, we conducted a quantitative comparison of our proposed model with state-of-the-art research on the CIFAR100 dataset as a benchmark. There are various input resolutions used in previous research, namely $32^2$ and $224^2$. The quantitative test results of our proposed model with the state-of-the-art are presented in Table \ref{tab:cifar100}. Based on our test results, SparseSwin is able to outperform previous research with an accuracy of 85.35\% with margins of $\pm 0.93\%$, $\pm 2.79\%$, $\pm 4.35\%$, $\pm 2.65\%$ and $\pm 1.75\%$ for ResNeXt-50 \cite{DBLP:journals/corr/XieGDTH16} , Transformer Local-Attention (NesT-B) \cite{zhang2022nested}, and CRATE-S \cite{yu2023white}, CRATE-B \cite{yu2023white}, CRATE-L \cite{yu2023white} respectively.

\begin{table}[bp]
\extracolsep{\fill}
\def\arraystretch{1.25}%
\centering
\large
\caption{Quantitative Comparison on CIFAR100}
\label{tab:cifar100}
\resizebox{\columnwidth}{!}{
    \begin{tabular}{c c c c c}
    \hline
     Model & \#Params(10 class) & Input Resolution & Model Type & Accuracy(\%) \\
    \hline
        ResNeXt-50\cite{DBLP:journals/corr/XieGDTH16} & 25.03 M & $224^2$ & Transformer & 84.42 \\
        NesT-B\cite{zhang2022nested} & 97.2 M & $32^2$ & Transformer & 82.56 \\
        CRATE-S\cite{yu2023white} & 13.12 M & $224^2$ & Transformer & 81.0 \\
        CRATE-B\cite{yu2023white} & 22.80 M & $224^2$ & Transformer & 82.7 \\
        CRATE-L\cite{yu2023white} & 77.64 M & $224^2$ & Transformer & 83.6 \\
    \hline
        SparseSwin & 17.58 M & $224^2$ & Transformer & \textbf{85.35} \\
    \hline
    \end{tabular}
}
\end{table}

\section{Conclusion}
\label{sec:conclusion}
In this paper, we propose a novel way to reduce the number of parameters and improve the capability of the Swin Transformer in image classification by using the Sparse Transformer Block (SparTa Block). This block consists of Sparse Token Converter and Regular Transformer Block to transform the feature representation obtained in Stage 3 of the Swin Transformer into a more concise representation while maintaining the important features required by the model to perform classification with the assistance of the attention mechanism in the regular transformer block. Through this modification, we obtained a smaller number of parameters compared to the Swin Transformer with 17.58 M parameters and obtained improvements in terms of accuracy on the ImageNet100, CIFAR10, and CIFAR100 datasets with accuracies of 86.96\%, 97.43\%, and 85.35\% respectively. Through this research, we hope that SparseSwin can perform well in classifying low-end devices and provide high performance in critical cases (e.g., biomedical analysis).

\bibliographystyle{splncs04}
\bibliography{refs} 

\end{document}